# Energy Management in Storage-Augmented, Grid-Connected Prosumer Buildings and Neighbourhoods Using a Modified Simulated Annealing Optimization

Rosemarie Velik, Pascal Nicolay

*Abstract*—This article introduces a modified simulated annealing optimization approach for automatically determining optimal energy management strategies in grid-connected, storage-augmented, photovoltaics-supplied prosumer buildings and neighbourhoods based on user-specific goals. For evaluating the modified simulated annealing optimizer, a number of test scenarios in the field of energy self-consumption maximization are defined and results are compared to a gradient descent and a total state space search approach. The benchmarking against these two reference methods demonstrates that the modified simulated annealing approach is able to find significantly better solutions than the gradient descent algorithm – being equal or very close to the global optimum – with significantly less computational effort and processing time than the total state space search approach.

## I. INTRODUCTION

OUR ELECTRICITY production and supply system, having in the last 50 years heavily built on electricity generated by centralized fossil fuel and nuclear power plants, is about to be transformed into a distributed electricity generation system consisting of smaller-scale renewable energy producers like buildings equipped with photovoltaics (and optionally also storage) systems [1], [2]. This massive structural change, accompanied by novel regulatory policies, brings along new challenges in terms of sustainable energy use [3], [4], [5]. In this context, the term "sustainable" can have different meanings depending on the concrete objective of the involved stakeholders. Examples for currently envisioned objectives are the maximization of the consumption of locally produced renewable energy, the achievement of energy autonomy, grid stability support, or a maximization of financial benefits [26], [27], [28]. Combinations of objectives are of course also possible. Each of these (sets of) objectives requires a particular "energy management strategy". Finding the optimal energy management strategy for each situation is however a task far from trivial and can massively benefit from computational assistance [6], [7], [8]. Optimization algorithms are an effective tool for identifying optimal strategies within complex energy management systems [9], [10].

In the context of building and mircrogrid energy management, several computational optimization approaches have already been proposed in literature for different applications. [11] presents an optimization approach for the effective energy management of a HVAC system using a metaheuristic simulation-evolutionary programming coupling method. [12] proposes a particle swarm optimization approach to optimize a control system having the task to improve user comfort and save energy. [13] aims to match load consumption from heating, ventilation, and air conditioning (HVAC) with available energy from a hybrid-renewable energy generation and energy storage system. A genetic-algorithm-based optimization approach together with a two-point estimate method is used to minimize the size of the photovoltaics and wind generation installation as well as the storage capacity to supply the HVAC load. [14] describes a dual evolutionary programming approach for a power system in which software agents co-evolve optimal operational behaviors for a simple microgrid configuration consisting of photovoltaics and conventional energy production sources, a battery storage, and partly controllable loads. [15] uses a genetic algorithm for optimizing the control of a stand-alone hybrid electrical system to achieve cost minimization over system lifetime. The electrical system can include renewable resources (e.g., wind, photovoltaics, hydro), batteries, a fuel cell, an AC generator and an electrolyzer.

In this article, we propose a modified simulated annealing approach for finding optimal control strategies for energy management in grid-connected, PV-supplied, storage augmented prosumer buildings and neighbourhoods in dependence of the objectives and goals of the involved stakeholders. To evaluate the performance of this approach, comparisons to a total state space search and a gradient descent method are provided for a range of different test scenarios aiming at an optimization of the local consumption of locally produced photovoltaics energy.

R. Velik and P. Nicolay are with CTR Carinthian Tech Research, Europastrasse 4/1, 9524 Villach, Austria (e-mail: first.last@ctr.at).

## II. Materials and Methods

*2.1 System Setup and Challenge Definition*

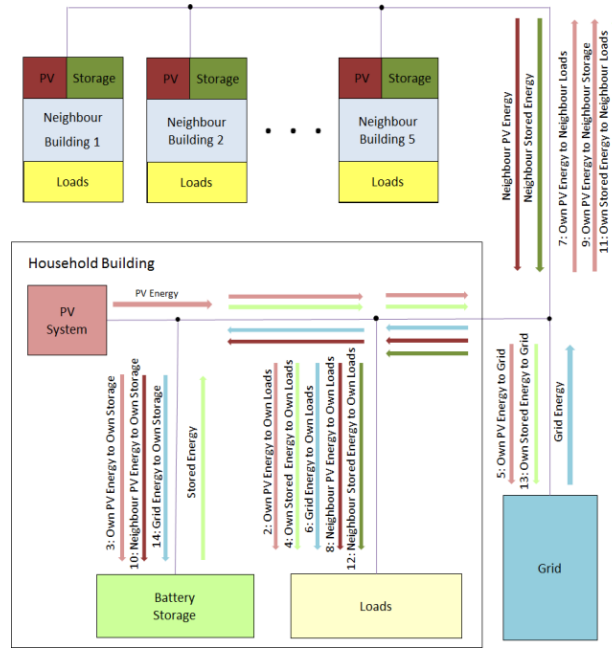

Fig. 1. Schematic overview of system topology of test setup including main building blocks and possible directions of energy exchange

Figure 1 gives an overview about the topology of the system for which an optimal energy management strategy shall be found. It illustrates a neighbourhood of six single-family houses in which each house consists of the following components: (1) a PV system, (2) a battery storage system, (3) household loads, (4) an interface to the neighbour buildings, (5) a grid connection.

The arrows between the building blocks indicate the principally possible directions of energy exchange within the system (in the further article referred to as possible "actions"), which are additionally summarized in Table I. Table I furthermore lists the pre-conditions for the execution of each action. In Section 2.4, different test scenarios will be specified and it will be indicated which of the energy exchange options will be supported in each scenario. The task of the energy optimizer developed in this article will be to prioritize/rank this list of supported actions based on the specified objectives of the energy management system (see Section 2.5.1). Optimization will be carried out in a distributed fashion, meaning that for each building, a separate instance of the optimizer is implemented.

The action with the ID 1 "Do No More Activity" means that all further actions ranked after this one will be ignored and thus not executed. This can be useful if the execution of certain principally supported actions worsens the optimization result. The actions with the IDs 2, 3, 4, 5, 6, 13, and 14 concern energy flows within a building and from the building to the grid and vice versa. The actions with the IDs 7, 8, 9, 10, 11, and 12 concern energy exchanges between always two specific buildings within the neighbourhood. To allow for such an energy exchange, the two corresponding buildings (the one that provides the energy and the one that consumes it) first have to negotiate and agree on this energy exchange (see also execution pre-conditions specified in Table I). The communication necessary for this negotiation takes place via specific communication channels between the buildings.

For our experiments, it was defined that for each action carried out, always the maximal possible amount of energy transfer is foreseen before the next action is considered. For the action "Own PV Energy to Own Loads", for instance, this would mean the following: If there is more own PV energy available than needed for the own loads, as much own PV energy is directed to the loads as necessary to cover their supply. Otherwise, if less own PV energy is available than needed for the own load supply, all available own PV energy is directed to the own loads. In analogy, for the action "Own PV Energy to Own Storage", this would mean: If more own PV energy is available than needed for completely charging the own storage, the storage is charged by the own PV energy until it is full. Otherwise, if less own PV energy is available than necessary for completely charging the own storage, all available own PV energy is used to charge the storage. In a similar way, this rule of "maximum possible energy transfer" is applied to all other actions.

TABLE I
SUPPORTED ACTIONS ACCORDING TO POSSIBLE DIRECTIONS OF ENERGY EXCHANGE BETWEEN SYSTEM COMPONENTS

| Action ID | Action Description | Execution Pre-Conditions |
|---|---|---|
| 1 | Do no more activity | None |
| 2 | Own PV energy to own loads | Own PV energy remaining & Own loads still need energy |
| 3 | Own PV energy to own storage | Own PV energy remaining & Own storage is not full |
| 4 | Own stored energy to own loads | Own storage is not empty & Own loads still need energy |
| 5 | Own PV energy to grid | Own PV energy remaining |
| 6 | Grid energy to own loads | Own loads still need energy |
| 7 | Own PV energy to neighbour loads | Own PV energy remaining & Neighbour loads still need energy & Mutual agreement for according energy exchange |
| 8 | Neighbour PV energy to own loads | Neighbour PV energy remaining & Own loads still need energy & Mutual agreement for according energy exchange |
| 9 | Own PV energy to neighbour storage | Own PV energy remaining & Neighbour storage is not full & Mutual agreement for according energy exchange |
| 10 | Neighbour PV energy to own storage | Neighbour PV energy remaining & Own storage is not full & Mutual agreement for according energy exchange |
| 11 | Own stored energy to neighbour loads | Own storage is not empty & Neighbour loads still need energy & Mutual agreement for according energy exchange |
| 12 | Neighbour stored energy to own loads | Neighbour storage is not empty & Own loads still need energy & Mutual agreement for according energy exchange |
| 13 | Own stored energy to grid | Own storage is not empty |
| 14 | Grid energy to own storage | Own storage is not full |

*2.2 Data Acquisition and Pre-Processing*

For developing and testing the energy optimizer, PV and load curves were acquired using a simple renewable energy data generator developed during the project "Vision Step I – Smart City Villach" [16], [17], [18], [19], [20]. This way, individual load profiles were generated for the 6 households for a period of 30 days with a resolution of one hour. Data were normalized to yield an average load consumption of 16kWh per household and day when averaging over the 30 days period.

Concerning the PV data acquisition, for each household, the same PV profile was generated for a period of 30 days with a resolution of one hour. Using the same PV profile for each building corresponds to the case that each building roof has the same orientation and tilt angle and the same number and type of PV modules. Such a situation is realistic in newly built residential estates realized by larger construction companies. Again, data were normalized to yield an average PV production of 16kWh per household and day when averaging over the 30 days. Accordingly, when averaging over the period of 30 days, the amount of produced PV energy within each building equaled the amount of energy consumed by its loads, which turned out to simplify data analysis.

TABLE II
SPECIFIED TEST SCENARIOS AND PRINCIPALLY POSSIBLE SELECTABLE ACTIONS FOR EACH OF THE SCENARIOS

| Scenario | 1: Do no more activity | 2: Own PV energy to own loads | 3: Own PV energy to own storage | 4: Own stored energy to own loads | 5: Own PV energy to grid | 6: Grid energy to own loads | 7: Own PV energy to neighbour loads | 8: Neighbour PV energy to own loads | 9: Own PV energy to neighbour storage | 10: Neighbour PV energy to own storage | 11: Own stored energy to neighbour loads | 12: Neighbour stored energy to own loads | 13: Own stored energy to grid | 14: Grid energy to own storage |
|---|---|---|---|---|---|---|---|---|---|---|---|---|---|---|
| 1: Individual PV self-consumption maximization without storage use | x | x | | | x | x | | | | | | | | |
| 2: Neighbourhood PV self-consumption maximization without storage use | x | x | | | x | x | x | x | | | | | | |
| 3a: Individual PV self-consumption maximization with individual storage use | x | x | x | x | x | x | | | | | | | | |
| 3b: Individual PV self-consumption maximization with individual storage use | x | x | x | x | x | x | | | | | | | x | x |
| 4a: Neighbourhood PV self-consumption maximization with individual storage use | x | x | x | x | x | x | x | x | | | | | | |
| 4b: Neighbourhood PV self-consumption maximization with individual storage use | x | x | x | x | x | x | x | x | | | | | x | x |
| 5a: Neighbourhood PV self-consumption maximization with neighbourhood storage use | x | x | x | x | x | x | x | x | x | x | x | x | | |
| 5b: Neighbourhood PV self-consumption maximization with neighbourhood storage use | x | x | x | x | x | x | x | x | x | x | x | x | x | x |

*2.3 System Modelling*

Figure 1 gave an overview about the topology of the energy system and its components, which was realized within a Visual Studio C++ simulation. To keep system modelling simple and to allow for an easier analysis of results, efficiency losses of individual components (PV modules, PV and battery inverters, batteries, wiring, etc.) were neglected for the current study. This way, analyses could be performed independent of the performance of currently available technologies, which are in fact changing and improving rapidly at present. This allowed for investigating the theoretically possible upper limits and potentials of the selected energy management strategies independent of any particular technology currently on the market.

In the specified test scenarios featuring energy storage (see Section 2.4), analyses were performed with battery storage capacities of 16 kWh per household. This storage size equaled the average amount of daily PV production and load consumption. The value of 16 kWh corresponded to the actually usable capacity, not to the nominal capacity labeled by storage producers. At the beginning of the 30 days test period, all storages were empty.

*2.4 Specified Test Scenarios*

To test the developed optimizers (see Section 2.5.2), five different test scenarios were specified from which the scenarios 3 to 5 were sub-divided into always two further sub-cases. The principal goal of each of the five scenarios was to optimize the local consumption of locally produced PV energy and thus to reduce energy exchange with the grid as far as possible (see Section 2.5.1 for further details). To achieve this objective, for each scenario, different actions can be selected (see Table II).

In the scenarios 1 and 3, for instance, each building aims at optimizing its local PV energy consumption individually while in the other cases, an energy exchange with neighbours is supported. In the scenarios 1 and 2, no buffering of PV energy in a battery storage is supported, while the other scenarios allow for such a temporal energy storage. The scenarios 3 to 4 only allow an individual storage use while scenario 5 allows the exchange of energy using the neighbour storages. Furthermore, the scenarios 3b, 4b, 5b support an interaction of the battery storage with the grid (charge storage from grid and feed stored energy into grid) while the other scenarios do not foresee this option.

TABLE III
SUB-OBJECTIVES FOR OPTIMIZATION AND THEIR ASSIGNED VARIABLES, TARGET VALUES AND PRIORITIES

| Sub-Objective | Optimization Variable | Target Value | Priority |
|---|---|---|---|
| All active loads need to be supplied with sufficient energy | Energy_Necessary_to_Obtain | 0 | 1 |
| No PV energy should remain unused | PV_Energy_Remaining | 0 | 2 |
| Direct consumption of PV energy produced within neighbourhood should be maximized | Direct_Local_PV_Consumption | MAX | 3 |
| Direct consumption of PV energy produced by own building should be maximized | Direct_Own_PV_Consumption | MAX | 4 |
| Stored PV energy should be used within the neighbourhood | Local_Storage_Consumption | MAX | 5 |
| Storing surplus PV energy within the own storage is preferred to storing surplus PV energy in the neighbour storages | Own_Storage_Loading | MAX | 6 |

*2.5 Optimizer*

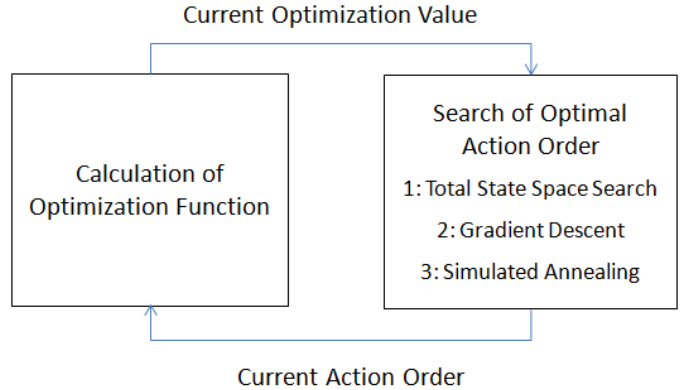

Fig. 2 Algorithm blocks and interfaces for finding optimal energy management strategies

To find the optimal energy management strategy, i.e. the best action order, for each of the scenarios listed in Table II, the optimizer was split into two blocks with defined interfaces (see Figure 2). Block 1 (depicted on the left) is responsible for the calculation of the optimization function based on the current action order provided by block 2. According to this current action order, the available energy is distributed within the system and a range of corresponding variables are calculated that keep track of the amount of energy directed to each component in each time instant. Based on the values of these variables, the optimization function is calculated (see Section 2.5.1 for details). The result of the optimization function is transmitted to block 2 (depicted on the right), which in further consequence suggests a new action order to block 1. Details about the internal functioning of block 2 are described in Section 2.5.2.1 for the case of the total state space search algorithm, in Section 2.5.2.2 for the gradient descent method, and in Section 2.5.2.3 for the simulated annealing approach.

*2.5.1 Calculation of Optimization Function*

The principal objective of each of the scenarios described in Table II is the maximization of the consumption of locally produced PV energy within the neighbourhood. To fulfill this principal objective, several sub-objectives have to be fulfilled which all have assigned a certain priority. Table III summarizes each of these sub-objectives, its corresponding optimization variable, and its "target value". The order of the list corresponds to the assigned priorities. Based on these optimization variables, the overall optimization value for the current action order is calculated in two parts (see Table IV).

Part 1: In accordance with the sub-objective prioritization given in Table III, the first two sub-objectives are translated into pre-conditions for calculating the optimization values of the other four sub-objectives. If a certain action order is NOT able to achieve that those first two optimization variables have the value zero in a certain time instant, the four remaining optimization variables will get assigned the value zero for this time instant (see part 1 of algorithm). This means that they will not contribute to the overall sum of each variable.

Part 2: As a second step, the four optimization variables calculated in step 1 are summed up using the weighted optimization function:

$$\text{Optimization Function} = 1/(w_1 \cdot Direct\_Local\_PV\_Consumption + w_2 \cdot Direct\_Own\_PV\_Consumption + w_3 \cdot Local\_Storage\_Consumption + w_4 \cdot Own\_Storage\_Loading)$$

The weights ($w_1$ to $w_4$) of this function correspond to the priority of each sub-objective as listed in Table III. For our simulations, a factor of always 10 was chosen between each weight (see Table IV, part 2), which proved to yield the desired results.

TABLE IV
PSEUDO-CODE FOR THE CALCULATION OF THE OPTIMIZATION FUNCTION

**Part 1: Calculation of Values of Individual Optimization Objectives**

```
Overall_Direct_Local_PV_Consumption=0;
Overall_Direct_Own_PV_Consumption=0;
Overall_Local_Storage_Consumption=0;
Overall_Own_Storage_Loading=0;
For each time instant i and each building j
    If((Energy_Necessary_to_Obtain[i][j]==0) AND
    (PV_Energy_Remaining[i][j]==0))
        Add the current values of the variables
                Direct_Local_PV_Consumption[i][j],
                Direct_Own_PV_Consumption[i][j],
                Local_Storage_Consumption[i][j],
                Own_Storage_Loading[i][j]
        to the variables
                Overall_Direct_Local_PV_Consumption,
                Overall_Direct_Own_PV_Consumption,
                Overall_Local_Storage_Consumption,
                Overall_Own_Storage_Loading.
    Else
        Add the value 0 to the variables
                Overall_Direct_Local_PV_Consumption,
                Overall_Direct_Own_PV_Consumption,
                Overall_Local_Storage_Consumption,
                Overall_Own_Storage_Loading.
    End
End
Calculate the average values per day and building of the
variables by dividing overall values through the number of
recorded days (30) and the number of buildings (6)
```

**Part 2: Calculation of Optimization Value of Overall Optimization Function**

```
Optimization_Function_Value=0;
w1=1000;
w2=100;
w3=10;
w4=1;

Optimization_Function_Value=1/(w1*Direct_Local_PV_Consumption+
w2*Direct_Own_PV_Consumption+w3*
Local_Storage_Consumption+w4*Own_Storage_Loading);
```

*2.5.2 Search of Optimal Action Order*

*2.5.2.1 Total State Space Search*

The total state space search approach determines the optimal energy management strategy for a certain set of objectives and possible actions by simply permuting through all possible action orders and calculating the corresponding optimization function value. For our study, this was realized using Alexander Bogomolyn's unordered permutation algorithm [21].

As all possibilities are explored by total state space search, it will always find the globally optimal solution and can thus serve as a reference method for benchmarking the performance of other approaches like our modified simulated annealing algorithm. However, the major disadvantage of the total state space search approach is its high computational complexity [22]. If the optimal action order for a set of x actions shall be found, the computational effort for this task is x factorial (x!) (see Table V). For larger sets of actions, exploring all possible permutations is thus no longer a feasible approach. To cope with the computational complexity of our test scenario 5b for instance, which consisted of 14 actions and thus required to investigate 87,178,000,000 different action combinations, processing had to be distributed over several machines and required a processing time of several months. It is obvious that this is no practicable solution for many applications.

*2.5.2.2 Gradient Descent*

The standard version of the gradient descent algorithm aims at minimizing a given optimization function by starting from an initial set of parameter values and then iteratively changing those values following the negative direction of the gradient of the function. For our application, this basic strategy was implemented as follows: The algorithm started witch a randomly chosen action order and calculated the optimization function for this action order. In subsequent steps, always two neighbouring positions in the action sequence were switched and the value of the corresponding optimization function was recalculated. The action order switch that yielded the largest negative change in the optimization value was taken into account for determining the next action sequence and a new round of "neighbor switching" was started. This process was repeated until a switching of neighbouring positions led to no further improvements in the optimization result.

While the gradient descent algorithm is relatively straightforward to implement and generally "inexpensive" in terms of processing time, its disadvantage is that it is prone to getting stuck in local minima.

*2.5.2.3  Simulated Annealing*

To overcome the "combinatorial explosion problem" of the total state space search algorithm and the gradient descent's tendency to get trapped in local minima, a modified version of the well-known standard simulated annealing (SA) algorithm [23] was implemented. The structure of the algorithm was based on the SA algorithm published in [24]. The SA algorithm was chosen for its ability to converge efficiently toward the optimum (i.e. maximum or minimum) of a given multivariable function in a large search space.

**Standard Simulated Annealing Approach**

The particular characteristic and strength of the SA it that it keeps exploring worse solutions throughout the whole optimization process. This means that it does not aim at systematically decreasing the value of a given multivariable function to find its absolute minimum. Instead, it occasionally accepts to degrade its current best solution to keep exploring other areas of the search space. This behavior helps avoiding being stuck in local minima, which is an especially useful property when a large search space, potentially full of local minima, has to be explored.

The decision of acceptance or refusal of a worse solution is based on the Metropolis criterion [23]. This criterion makes the acceptance of the solution more likely if the 'temperature' is high and the degradation of the solution is small. The term 'temperature' as used here refers to a parameter that steadily decreases with time over the whole optimization process. It is chosen as an analogy to the physical process of thermal annealing, which is mimicked by the SA algorithm. With the 'temperature' decreasing over time, also the probability for accepting worse solutions is starting to decrease.

Once the temperature gets below a certain threshold value, the SA algorithm stops exploring further worse solutions and keeps converging steadily toward the closest minimum in the search space. Accordingly, at the end of the optimization process, the SA algorithm behaves like a gradient descent optimizer [24]. Optimization finally stops if the value to be optimized cannot be further minimized.

The exploration of the search space itself proceeds in the following way: For each new iteration cycle, the SA algorithm generates a new value for each of the multivariable function's variables. These new values are calculated by adding a random quantity to the reference values (or current best solution) of the given variables. The random quantity is usually obtained using a Gaussian random number generator, whose standard deviation (SD) is adjusted beforehand according to the desired exploration range. The set of new values (i.e. the new solution) is then used to compute the multivariable function. If the result is better than the one obtained in the previous iteration, the new solution is kept as the new reference and a new cycle begins. If the result is worse, the Metropolis criterion is applied to accept or refuse the new solution. If accepted, the new solution becomes the new reference and a new cycle begins.

**Modified Simulated Annealing Approach**

The aim of our article was to use the SA method for rearranging action orders until the best possible ranking (i.e. prioritization) of the actions specified in Table II is achieved for a particular scenario according to its set objective(s). For this purpose, some adaptations had to be made to the standard SA algorithm.

First, it was necessary to convert the ranking of an action into a value of a variable, which could change continuously in a given range.

Second, it had to be assured that two actions did not occupy the same position in the ranking. This means that the two corresponding variables cannot have the same value.

These two modifications were solved in the following way: Each action got assigned an independent position variable of the data type double. For each variable, the same variation domain was specified equaling the number of selectable actions. For instance, if ten actions were used, the positions of the actions would be given by ten position variables located on a continuous zero-to-ten scale. Initially, the value of the position variables was determined randomly. In consecutive optimization steps, they were altered using a Gaussian random number generator with a specified standard deviation. To avoid that two position variables have exactly the same value, they were checked for equality and altered again if necessary.

The following example shall contribute to the clarification of the working principle of our modified simulated annealing algorithm: If for instance three actions have to be ranked, a first ranking is generated by assigning random values between 0 and 3 to each action. This serves as a starting point for the optimization. A possible example for this would be:

action 1: position variable=1.58

    action 2: position variable=2.25

    action 3: position variable=0.57

which would result in the following action ranking:

    action 2 – action 1 – action 3

This first ranking would be injected into the "Calculation of Optimization Function" block (see Figure 2) to get a first result for the optimization function. The SA algorithm would subsequently alter the values of the three position variables to achieve a minimization of the optimization function, yielding the best possible ranking of actions. For instance, the SA algorithm would yield the following values at the end of the optimization:

    action 1: position variable=0.95

    action 2: position variable=2.15

    action 3: position variable=2.99

which would result in the following action ranking:

    action 3 – action 2 – action 1

It has to be stressed that in our modified version of the SA algorithm, the variables are not directly used for the computation of the optimization function. They are first translated into a ranking list of actions. Based on this ranking, certain optimization variables forming part of the optimization function are calculated (see Section 2.5.1). This means that different sets of position variable values can lead to exactly the same ranking result. For instance, the following set of positions:

    action 1: position variable=0.52

    action 2: position variable=1.56

    action 3: position variable=2.05

would also result in the action ranking:

    action 3 – action 2 – action 1

Actually, the optimization error only changes in case that ranking positions get swapped. To ensure that the modified SA algorithm does not get stuck in a given non-optimal ranking, this swapping has to be ensured to occur frequently (at least at the beginning of the optimization). To enable frequent swapping, the standard deviation SD has to be large enough and thus has to be adjusted prior to the optimization. In our case, we used a SD of 1.5. This value was the result of a trial and error procedure, conducted over a great number of test optimizations involving different numbers of actions. The theoretical derivation of the optimum SD for this kind of optimization problem will be discussed in future work.

### III. Results and Discussion

This section is concerned with illustrating and comparing the results of the modified simulated annealing approach with the total space state search and gradient descent reference methods. For this purpose, Section 3.1 presents the computational effort of these algorithms for the different test scenarios. Section 3.2 comprises a comparison of the found action orders and corresponding optimization values of the algorithms. Section 3.3 illustrates the achieved local PV energy consumption based on the energy management strategies found by the algorithms.

#### 3.1 Computational Effort of Algorithm

Table V compares the computational effort of the total state space search algorithm, the gradient descent algorithm, and the modified simulated annealing approach for the different investigated scenarios. As can be seen from Table V, the gradient descent algorithm requires very little computation effort. However, as illustrated in Section 3.2, it shows the drawback of getting stuck in local minima for most of the test scenarios. Accordingly, it proved to be only of limited use for the given application. The total state space search approach works well for small sets of selectable actions. However, for larger sets, it leads to a "combinatorial explosion". Accordingly, for finding the best prioritization of 10 or more actions, the modified simulated annealing approach outperformed the total state space search algorithm in terms of computational effort. For the largest investigated action set (scenario 5b with 14 possible actions), the computational effort of the state space search algorithm was 14 factorial (14!), which was 92,458 times higher than the effort of the modified simulated annealing approach.

TABLE V
Comparison of Computational Effort of Total State Space Search, Gradient Descent, and Modified Simulated Annealing

| Scenario ID | Number of Selectable Actions | Number of Iterations | | |
|:---:|:---:|:---:|:---:|:---:|
| | | Total State Space Search | Gradient Descent | Modified Simulated Annealing |
| 1 | 4 | 24 | 5 | 269400 |
| 2 | 6 | 720 | 13 | 404100 |
| 3a | 6 | 720 | 13 | 404100 |
| 3b | 8 | 40320 | 17 | 538800 |
| 4a | 8 | 40320 | 17 | 538800 |
| 4b | 10 | 3628800 | 21 | 673500 |
| 5a | 12 | 479001600 | 25 | 808200 |
| 5b | 14 | 8.7178E+10 | 29 | 942900 |

*3.2 Determined Optimal Action Orders and Corresponding Optimization Values*

TABLE VI
ACTION ORDERS DETERMINED BY OPTIMIZATION ALGORITHMS

| Scenario ID | Action Orders | | |
|---|---|---|---|
| | Total State Space Search | Gradient Descent | Modified Simulated Annealing |
| 1 | 2 5 6 1<br>2 6 5 1 | 2 5 6 1 | 2 6 5 1 |
| 2 | 2 7 8 6 5 1<br>2 8 7 5 6 1<br>2 7 5 8 6 1 | 2 7 5 8 6 1 | 2 7 8 6 5 1 |
| 3a | 2 3 4 6 5 1<br>2 4 3 6 5 1<br>2 4 6 3 5 1 | 2 5 4 3 6 1 | 2 4 3 6 5 1 |
| 3b | 2 3 4 5 6 1 X X<br>2 4 3 6 5 1 X X<br>2 4 6 3 5 1 X X | 2 5 14 3 6 1 X X | 2 3 4 5 6 1 X X |
| 4a | 2 7 8 4 3 6 5 1<br>2 7 3 5 8 4 3 1<br>2 7 3 5 8 4 6 1 | 2 7 4 5 6 1 X X | 2 7 8 4 3 6 5 1 |
| 4b | 2 8 7 4 3 6 5 1 X X<br>2 7 3 5 8 4 3 1 X X<br>2 8 7 3 5 4 6 1 X X | 14 2 13 3 6 1 X X X X | 2 8 7 4 3 6 5 1 X X |
| 5a | 2 7 8 3 4 9 10 11 12 5 6 1<br>2 8 7 3 4 9 10 12 11 5 6 1<br>2 7 8 4 3 12 11 10 9 6 5 1 | 2 11 10 3 1 X X X X X X X | 2 7 8 4 3 6 12 11 10 9 5 1<br>(closest match) |
| 5b | 2 7 8 3 4 9 10 11 12 5 6 1 X X<br>2 8 7 3 4 10 9 11 12 6 5 1 X X<br>2 8 7 4 3 12 11 10 9 6 5 1 X X | 13 2 10 3 1 X X X X X X X X | 2 8 7 4 3 6 12 11 10 9 5 1 X X<br>(closest match) |

TABLE VII
VALUES OF INDIVIDUAL OPTIMIZATION VARIABLES OF SUB-GOALS AND VALUE OF OVERALL OPTIMIZATION FUNCTION

| | Total State Space Search | | | | | | | Gradient Descent | | | | | | | Modified Simulated Annealing | | | | | | |
|---|---|---|---|---|---|---|---|---|---|---|---|---|---|---|---|---|---|---|---|---|---|
| Scenario ID | Not provided load supply [kWh] | PV energy waste [kWh] | Direct local PV energy consumption [kWh] | Direct own PV energy consumption [kWh] | Local storage consumption [kWh] | Own storage loading [kWh] | Optimization value$^{-1}$ | Not provided load supply [kWh] | PV energy waste [kWh] | Direct local PV energy consumption [kWh] | Direct own PV energy consumption [kWh] | Local storage consumption [kWh] | Own storage loading [kWh] | Optimization value$^{-1}$ | Not provided load supply [kWh] | PV energy waste [kWh] | Direct local PV energy consumption [kWh] | Direct own PV energy consumption [kWh] | Local storage consumption [kWh] | Own storage loading [kWh] | Optimization value$^{-1}$ |
| 1 | 0 | 0 | 5.5 | 5.5 | 0 | 0 | **6016** | 0 | 0 | 5.5 | 5.5 | 0 | 0 | **6016** | 0 | 0 | 5.5 | 5.5 | 0 | 0 | **6016** |
| 2 | 0 | 0 | 5.8 | 5.5 | 0 | 0 | **6315** | 0 | 0 | 5.8 | 5.5 | 0 | 0 | **6315** | 0 | 0 | 5.8 | 5.5 | 0 | 0 | **6315** |
| 3a | 0 | 0 | 5.5 | 5.5 | 8.8 | 9 | **6113** | 0 | 0 | 5.5 | 5.5 | 0 | 0 | **6017** | 0 | 0 | 5.5 | 5.5 | 8.8 | 9 | **6113** |
| 3b | 0 | 0 | 5.5 | 5.5 | 8.8 | 9 | **6113** | 0 | 0 | 5.5 | 5.5 | 0 | 0 | **6017** | 0 | 0 | 5.5 | 5.5 | 8.8 | 9 | **6113** |
| 4a | 0 | 0 | 5.8 | 5.5 | 8.6 | 8.7 | **6409** | 0 | 0 | 5.8 | 5.5 | 0 | 0.5 | **6315** | 0 | 0 | 5.8 | 5.5 | 8.6 | 8.7 | **6409** |
| 4b | 0 | 0 | 5.8 | 5.5 | 8.6 | 8.7 | **6409** | 0 | 0 | 5.5 | 5.5 | 0 | 10.5 | **6027** | 0 | 0 | 5.8 | 5.5 | 8.6 | 8.7 | **6409** |
| 5a | 0 | 0 | 5.8 | 5.5 | 8.8 | 8.7 | **6411** | 0 | 0 | 4.9 | 4.9 | 9 | 8.7 | **5471** | 0 | 0 | 5.8 | 5.5 | 8.6 | 8.7 | **6409** |
| 5b | 0 | 0 | 5.8 | 5.5 | 8.8 | 8.7 | **6411** | 0 | 0 | 4.1 | 4.1 | 0 | 10.5 | **4541** | 0 | 0 | 5.8 | 5.5 | 8.6 | 8.7 | **6409** |

The tables VI and VII summarize the determined optimal action orders and corresponding optimization values for the analyzed algorithms. Actions that were prioritized behind the action "Do No More Activity" are not listed in Table VI as they are irrelevant. Instead their position is marked with an 'X'.

By running the total state space search optimizer, it could be demonstrated that for the investigated scenarios, more than one action combination can lead to the same optimal solution. Using the total state space search algorithm, all action combinations that led to this optimal solution were annotated. In Table VI, several of these optimal solutions are listed for each scenario exemplarily.

As can be seen from Table VI and VII, the gradient descent algorithm only found a globally optimal solution for scenario 1 and scenario 2. In contrast to this, the modified simulated annealing approach was in all but scenario 5 able to find a globally optimal solution. For scenario 5a and 5b, a solution was found that was very close to the global optimum. Only one action (Grid Energy to Own Loads) was prioritized in a not completely optimal order. Concerning the optimization function, this corresponded to an optimization error of 0.03 %. Concerning the scenario's objective of local PV energy consumption maximization, this resulted in 0.19 kWh (or 1.3 %) less local PV energy consumption per household and day than in case of the global optimum (see Section 3.3).

### 3.3 Achieved Local PV Energy Self Consumption

Using the action orders determined with the total state space search algorithm, the gradient descent algorithm, and the modified simulated annealing algorithm, the total local PV energy consumption within the neighbourhood was determined as well as the components of which it was made up (see Table VIII and figures 3 to 5).

As can be derived from these figures and table, for the scenarios 1 and 2, the same results could be achieved by all three algorithms. For the scenarios 3 and 4, only the total state space search and the modified simulated annealing algorithm found the same optimal solution. As described in Section 3.2, in scenario 5, the total local PV energy consumption within the neighbourhood was slightly lower for the modified simulated annealing solution than for the total state space solution due to a non-optimal prioritization of the action "Grid Energy to Own Loads". This had the consequence that no stored PV energy was exchanged with the neighbours, which resulted in 1.3 % less local use of stored PV energy within the neighbourhood.

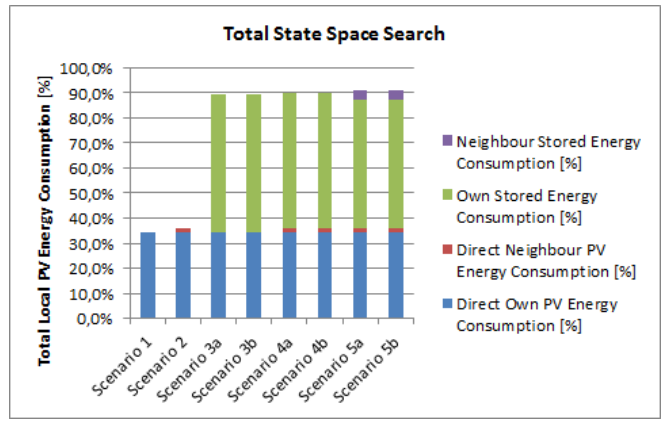

Fig. 3  Average local PV energy consumption and contributing "sources" per household and day achieved by total state space search optimizer

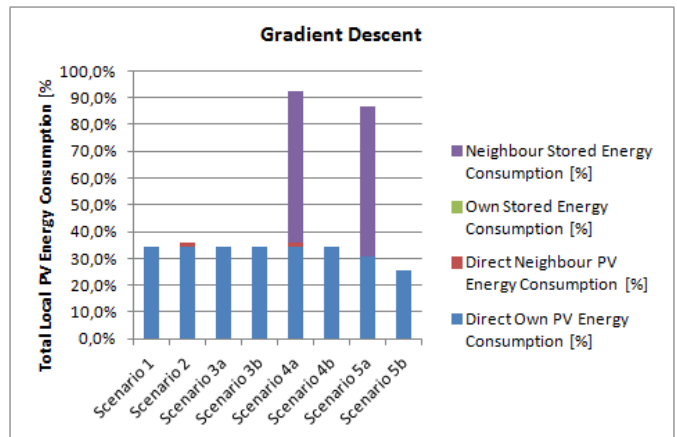

Fig. 4  Average local PV energy consumption and contributing "sources" per household and day achieved by modified gradient descent optimizer

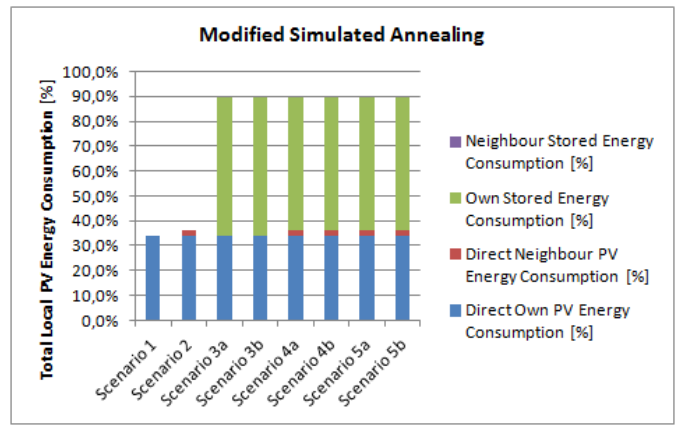

Fig. 5 Average local PV energy consumption and contributing "sources" per household and day achieved by modified simulated annealing optimizer

TABLE VIII
AVERAGE ACHIEVED LOCAL PV ENERGY CONSUMPTION PER HOUSEHOLD AND DAY IN % AND "SOURCES" CONTRIBUTING TO IT

| Scenario ID | Total State Space Search | | | | | Gradient Descent | | | | | Modified Simulated Annealing | | | | |
|---|---|---|---|---|---|---|---|---|---|---|---|---|---|---|---|
| | Direct own PV energy consumption [%] | Direct neighbour PV energy consumption [%] | Own stored energy consumption [%] | Neighbour stored energy consumption [%] | Total local PV energy consumption [%] | Direct own PV energy consumption [%] | Direct neighbour PV energy consumption [%] | Own stored energy consumption [%] | Neighbour stored energy consumption [%] | Total local PV energy consumption [%] | Direct own PV energy consumption [%] | Direct neighbour PV energy consumption [%] | Own stored energy consumption [%] | Neighbour stored energy consumption [%] | Total local PV energy consumption [%] |
| 1 | 34.2 | 0.0 | 0.0 | 0.0 | **34.2** | 34.2 | 0.0 | 0.0 | 0.0 | **34.2** | 34.2 | 0.0 | 0.0 | 0.0 | **34.2** |
| 2 | 34.2 | 1.9 | 0.0 | 0.0 | **36.0** | 34.2 | 1.9 | 0.0 | 0.0 | **36.0** | 34.2 | 1.9 | 0.0 | 0.0 | **36.0** |
| 3a | 34.2 | 0.0 | 55.1 | 0.0 | **89.3** | 34.2 | 0.0 | 0.0 | 0.0 | **34.2** | 34.2 | 0.0 | 55.1 | 0.0 | **89.3** |
| 3b | 34.2 | 0.0 | 55.1 | 0.0 | **89.3** | 34.2 | 0.0 | 0.0 | 0.0 | **34.2** | 34.2 | 0.0 | 55.1 | 0.0 | **89.3** |
| 4a | 34.2 | 1.9 | 53.6 | 0.0 | **89.7** | 34.2 | 1.9 | 0.0 | 56.3 | **90.4** | 34.2 | 1.9 | 53.6 | 0.0 | **89.7** |
| 4b | 34.2 | 1.9 | 53.6 | 0.0 | **89.7** | 34.2 | 0.0 | 0.0 | 0.0 | **34.2** | 34.2 | 1.9 | 53.6 | 0.0 | **89.7** |
| 5a | 34.2 | 1.9 | 51.1 | 3.7 | **90.9** | 30.5 | 0.0 | 0.0 | 56.3 | **86.8** | 34.2 | 1.9 | 53.6 | 0.0 | **89.7** |
| 5b | 34.2 | 1.9 | 51.1 | 3.7 | **90.9** | 25.8 | 0.0 | 0.0 | 0.0 | **25.8** | 34.2 | 1.9 | 53.6 | 0.0 | **89.7** |

## 4 CONCLUSION

This article presented a modified simulated annealing optimizer for finding optimal energy management strategies in storage-augmented grid-connected renewable prosumer buildings and neighbourhoods. Optimization objective was the maximization of the local consumption of locally produced PV energy. The results of the modified simulated annealing optimizer were evaluated by comparing them against the globally optimal solution determined by a computationally expensive total state space search approach searching through all possible action combinations and the solution obtained by a computationally inexpensive gradient descent algorithm. While the results of the gradient descent algorithm demonstrated the necessity of developing a method not getting easily trapped in local minima, the modified simulated annealing approach found the global optimum in six out of the eight tested scenarios. For the remaining two scenarios, it found a local optimum that was very close to the global optimum (0.03 % error of the optimization function, which translated into 1.3 % less local consumption of locally produced PV energy).

For complex scenarios involving many different possible actions, the modified simulated annealing approach proved to be computationally much less expensive than the total state space search method. For the two scenarios with the small remaining optimization error, the computational effort was up to 92,458 times less than the effort for calculating the global optima with the total state space search approach.

This demonstrated that the modified simulated annealing approach is a very powerful approach for finding optimal energy management strategies in large search spaces. Further work on this issue targeting also other energy management applications will be addressed in future work. Apart from this, similar as for the modified simulated annealing approach introduced here, the possibility will be explored to also modify other optimization algorithms (e.g. GRASP) to enable them to map action orders to optimization values.

ACKNOWLEDGMENT

The work reported in this article has been co-funded by the European Commission within the INTERREG Program for supporting small and medium sized companies (SME) in Italy and Carinthia (Austria) and the Austrian Research Promotion Agency (FFG) within the program Fit4Set (project Vision Step I).